\documentclass[a4paper,twoside]{article}

\usepackage{epsfig}
\usepackage{subcaption}
\usepackage{calc}
\usepackage{amssymb}
\usepackage{amstext}
\usepackage{amsmath}
\usepackage{amsthm}
\usepackage{multicol}
\usepackage{pslatex}
\usepackage{apalike}
\usepackage{SCITEPRESS}     
\usepackage{comment}

\begin{document}

\title{Generating Adequate Distractors for Multiple-Choice Questions}

\author{\authorname{Cheng Zhang\sup{1}, Yicheng Sun\sup{1}, Hejia Chen\sup{2} and Jie Wang\sup{1}}
\affiliation{\sup{1}Department of Computer Science, University of Massachusetts, Lowell,  MA 01854, USA}
\affiliation{\sup{2}School of Computer Science and Technology, Xidian University, Xi'an 710126, PRC}
\email{\{Cheng\_Zhang, Yicheng\_Sun\}@student.uml.edu, chenhj2000@stu.xidian.edu.cn, wang@cs.uml.edu}
}


\abstract{This paper presents a novel approach to automatic generation of adequate distractors for a given question-answer pair (QAP) generated from a given article to form an adequate multiple-choice question (MCQ). Our method is a combination of part-of-speech tagging, named-entity tagging, semantic-role labeling, regular expressions,
domain knowledge bases, word embeddings, word edit distance, WordNet, and other algorithms.
We use the US SAT (Scholastic Assessment Test) practice reading tests as a dataset to produce QAPs and generate three distractors for each QAP to form an MCQ. We show that, via experiments and evaluations by human judges, each MCQ has at least one adequate distractor and 84\% of MCQs have three adequate distractors.}

\onecolumn \maketitle \normalsize \setcounter{footnote}{0} \vfill

\section{\uppercase{Introduction}}
\noindent
Generating MCQs on a given article is a quick and effective method for assessing the reader's comprehension
of the article. An MCQ  typically consists of a QAP and a few distractors. This paper
is focused on generating adequate distractors for a given QAP in connection to the
underlying article. 
%
%

An adequate distractor must satisfy the following requirements:
(1) it is an incorrect answer to the question;
(2) it is grammatically correct and consistent with the underlying article;
(3) it is semantically related to the correct answer; and
(4) it provides distraction so that the correct answer could be identified only with some understanding of the underlying article.

Given a QAP for a given article,
we study how to generate adequate distractors that are grammatically correct and semantically related to the correct answer in the sense that the distractors, while incorrect, look
similar to the correct answer with a sufficient distracting effect---that is, it should be hard to
distinguish distractors from the correct answer without some understanding of the underlying article.
A distractors could be a single word, a phrase, a sentence segment, or a complete sentence.

In particular, we are to  generate three adequate 
distractors for a QAP to form an MCQ. One way to generate a distractor is to substitute a word or a phrase contained in the answer with an appropriate word or a phrase that maintains the original part of speech. 
Such a word or phrase could be an answer itself or contained in an answer sentence or sentence segment. For convenience, we refer to such a word or phrase as a target word. 

If a target word is a number with an explicit or implicit quantifier,
or anything that can be converted to a
number,
we call it a type-1 target.
If a target word is a person, location, or organization,
we call it a type-2 target. 
Other target words (nouns, phrasal nouns, adjectives, verbs, and adjectives) are referred to as type-3 targets.
We use different methods to generate distractors for targets of different types.

Our distractor generation method is a combination of part-of-speech (POS) tagging \cite{toutanova2003feature}, 
named-entity (NE) tagging \cite{nadeau2007survey,ali2010automation,peters2017semi}, 
semantic-role labeling \cite{martha2005proposition,shi2019simple},
regular expressions, domain knowledge bases on people, locations, and organizations,
word embeddings (such as
Word2vec \cite{10.5555/2999792.2999959}, GloVe \cite{pennington-etal-2014-glove}, Subwords \cite{bojanowski2017enriching}, 
 and
spherical text embedding \cite{sphericalembedding2019}), word edit distance \cite{levenshtein1966binary}, WordNet (https://wordnet.princeton.edu), and some other algorithms.
We show that, via experiments, our method can generate adequate distractors for a QAP to form an MCQ with
a high successful rate. 

The rest of the paper is organized as follows: We describe related work in
Section \ref{sec:relatedwork} and present our distractior generation method in Section \ref{sec:method}.
We then show in Section \ref{sec:evaluation} that,
via experiments on the official SAT practice reading tests and evaluation by human judges,
each MCQ has at least one adequate distractor and 84\% of MCQs have three adequate distractors.
We conclude the paper in Section \ref{sec:conclusion}.

\section{\uppercase{Related Work}} \label{sec:relatedwork}

\noindent
Methods of generating adequate distractors for MCQs are typically following 
two directions: domain-specific knowledge bases 
and semantic similarity \cite{pho-etal-2014-multiple,CH2018Automatic}.

Methods in the first direction are focused on lexical information 
have used part-of-speech(POS) tags, word frequency, WordNet, domain ontology, distributional hypothesis, 
and pattern matching, to find the target word's synonym, hyponym and hypernym as distractor candidates. \cite{10.3115/1118894.1118897,10.1007/978-3-642-28885-2_19,Susanti2015AutomaticGO}

Methods in the second direction 
analyze the semantic similarity of the target word using Word2Vec model for generating distractor candidates \cite{Jiang2017DistractorGF,Susanti2018AutomaticDG}. 

However,
it is difficult to use Word2Vec or other word-embedding methods to find adequate 
distractors for polysemous answer words.
Moreover, previous efforts have focused on finding some forms of distractors, instead of making them look more distracting. This paper is an attempt to tackle these issues.

\section{\uppercase{Distractor Generation}} \label{sec:method}

\begin{figure}[!h]
  \centering
  {\epsfig{file = 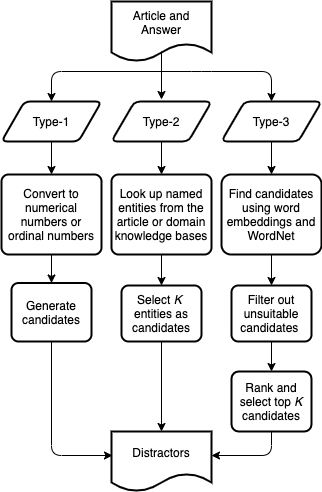, width = 2.8in}}
  \caption{Distractor generation flowchart}
  \label{fig:dgf}
 \end{figure}

\noindent
Our method takes the original article and the answer as input, and generates distractors as output. Figure~\ref{fig:dgf} depicts the data flow of our method.

Answers in QAPs are classified into two kinds. The first kind consists of just a single target word while the second kind consists of multiple target words. The latter is the case when  the
answer is a sentence or a sentence segment. 

For an answer of the first kind, if it is a type-1 or type-2 target, we use the methods
described in Section \ref{sec:2} to generate three distractors; if it is a type-3 target, we use
the method described in Section \ref{sec:3} to generate distractor candidates.
If there are at least three candidates, then select three candidates
with the highest ranking scores. 

For an answer of the second kind, for each type of a target word contained in it, we use the methods described in both Sections \ref{sec:2} and
\ref{sec:3} to generate distractors for target words in a fixed ordered preference of subjects, objects, adjectives for subjects, adjectives for objects,
predicates, adverbs, which can be obtained by semantic-role labeling.
Target words are replaced
according to the following preference:
%
type-1 temporal, type-1 numerical value, type-2 person, type-2 location,
type-2 organization, type-3 noun (phrasal noun), type-3 adjective, type-3 verb (phrasal verb), and type-3 adverb. 

If the number of distractors for a given preference is less than three, then we generate extra distractors for a target word in the next preference. If all preference is gone through we still need more distractors, we could extend the selection threshold values to allow more candidates to be selected.

\subsection{Distractors for Type-1 and Type-2 Targets}
\label{sec:2}

If a type-1 target is a point in time, a time range, a numerical number, an ordinal number, or anything that can be converted to a numerical number or an ordinal number (e.g. Friday may be converted to 5), which can be
recognized by regular expressions based on a POS tagger, then we devise several algorithms to alter time and number, and randomly select one of these algorithms when generating distractors. For example, we may  increase or decrease the answer value by one or two units, change the answer value at random from a small range of values around the answer,
or simply change the answer value at random. If a numerical value or an ordinal number
is converted from a word, then the result is converted back to the same form. For example,
suppose that the target word is ``Friday", which is converted to a number 5. If the distractor
is a number 4, then it is converted to Thursday. 

If a type-2 target is a person, then we first look for different person names that appear in the article using an NE tagger to identify them, and then randomly choose a name as a distractor. 
If there are no other names in the article, 
then we use Synonyms (http://www.synonyms.com) or a domain knowledge base on notable people we constructed to find a distractor. 
If a type-2 target is a location or an organization, we find a distractor in the same way by first
looking for other locations or organizations in the article, and then using
Synonyns and domain knowledge bases to look for them if they cannot be found in
the article.
For example, If the target word is a city, then a distractor should also  be city that is "closely" related to the target word. Distractors to the answer word "New York" should be cities in the same league, such as ``Boston", ``Philadelphia", and ``Chicago".  

\subsection{Distractors for Type-3 Targets}
\label{sec:3}
\noindent
For a type-3 target, 
we find distractor candidates 
using word embeddings 
with similarity in a threshold interval (e.g.,[0.6,0.85]) so that
a candidate is not too close nor too different from the correct answer
and 
hypernyms using WordNet \cite{10.1145/219717.219748}. 
Note that a similarity interval of $[0.6,0.85]$ for word embeddings often
include antonyms of the target word, and we can use WordNet or an online dictionary to determine
antonyms.

Not all distractor candidates are suitable. Thus, we first filer out unsuitable candidates as follows:
\begin{enumerate}
\item Remove distractor candidates that contain the target word, for it may be
too close to the correct answer. For example, if 
``breaking news" is
a generated distractor candidate for the target word "news", then it is
removed from the candidate list since it contains the target word.

\item Remove distractor candidates that have the same prefix of the target word with edit distance less than three, for such candidates may often be misspelled words from the target word.
For example, suppose the target word is "knowledge", then 
Word2vec may return a misspelled candidate "knowladge" with a high similarity,
which should be removed.

\end{enumerate}

We then rank each remaining candidate using the following measure:
\begin{enumerate}
\item Compute the Word2vec cosine similarity score $S_v$ for each distractor candidate $w_c$ with the 
target word $w_t$. Namely, $$S_v = \mbox{sim}(v(w_c),v(w_t)),$$
where $v(w)$ denotes a word embedding of $w$.

\item Compute the WordNet WUP score \cite{10.3115/981732.981751} $S_n$ for each distractor candidate with the target word. If the distractor candidate cannot be found in the WordNet dataset, set the WUP score to 0.1
for the following reason: If a word with a high score of word-embedding similarity to the target word but does not exist in the WordNet dataset, then it is highly likely that the word is misspelled, and so its ranking score should be reduced.

\item Compute the edit distance score $S_d$ of each distractor candidate with target word by the
following formula:
\begin{equation*}\label{eq1}
    S_d = 1 - \frac{1}{1 + e^E},
\end{equation*}
where $E$ is the edit distance. Thus, a lager edit distance $E$ results in a smaller score $S_d$.

\item Compute the final ranking score $R$ for each distractor candidate $w_c$ with respect to the target word $w_t$ by
\begin{align*}\label{eq2}
R'(w_c,w_t) &= \left\{
\begin{array}{ll}
\frac{1}{4}(2S_v + S_n + S_d), &\mbox{if $w_c$ is an} \\
							&\mbox{antonym of $w_t$}, \\
\frac{1}{3}(S_v + S_n+S_d), & \mbox{otherwise},
\end{array}\right. \\
R(w_c,w_t) &=  - R'(w_c,w_t)\log R'(w_c,w_t).\\
\end{align*}
Note that $S_v, S_n, S_d$ are each between 0 and 1, and so
$R'(w_c,w_t)$ is between 0 and 1, which implies that $- \log R'(w_c,w_t) > 0$.
Also note that we give more weight to antonyms.

\end{enumerate}


\section{\uppercase{Evaluations}}
\label{sec:evaluation}
\noindent
We implemented our method using the latest versions of POS tagger\footnote{https://nlp.stanford.edu/software/tagger.shtml}, NE tagger \cite{peters2017semi},
semantic-role labeling \cite{shi2019simple}, and fastText \cite{mikolov2018advances}.
We used the US SAT  practice reading tests\footnote{https://collegereadiness.collegeboard.org/sat/practice/full-length-practice-tests} as a dataset for evaluations. 
There are a total of eight SAT practice reading tests, each consisting of five articles for a total of 40 articles.
Each article in the SAT practice reading tests consists of around 25 sentences and we
generated about 10 QAPs from each article. To evaluate our distractor generation algorithm, we selected independently at random slightly over 100 QAPs. After removing a smaller number of QAPs with
pronouns as target words, we have a total of 101 QAPs for evaluations.

We generated 3 distractors for each QAP for a total of 303 distractors, and evaluated distractors 
based on the following criteria:
\begin{enumerate}
\item A distractor is \textsl{adequate} if it is grammatically correct and relevant to the question with distracting effects.
\item An MCQ is \textsl{adequate} if each of the three distractors is adequate.
\item An MCQ is \textsl{acceptable} if one or two  distractors are adequate. 
\end{enumerate}
We define two levels of distracting effects: (1) \textsl{sufficient distraction}: It requires 
an understanding of the underlying article to choose the correct answer;
(2) \textsl{distraction}: It only requires an understanding of the underlying question to choose
the correct answer.
A distractor has no distracting effect if it can be determined wrong by just looking at
the distractor itself.

Evaluations were carried out by humans and the results are listed below:
\begin{enumerate}
\item All distractors generated by our method are grammatically correct.
\item 98\% distractors (296 out of 303) are relevant to the QAP with distraction.
\item 96\% distractors (291 out of 303) provide sufficient distraction.
\item 84\% MCQs are adequate.
\item All MCQs are acceptable (i.e., with at least one adequate distractor).
\end{enumerate}

Given below are a few adequate MCQs with automatically generated distractors by our method:

\subsubsection*{Example 1}
Question: What does no man like to acknowledge? (SAT practice test 2 article 1)

Correct answer: that he has made a mistake in the choice of his profession.

Distractors:
\begin{enumerate}
\item that he has made a mistake in the choice of his association.
\item that he has made a mistake in the choice of his engineering.
\item that he has made a mistake in the way of his profession.
\end{enumerate}

\subsubsection*{Example 2}

When should ethics apply? (SAT practice test 2 article 2)

Correct answer: 
when someone makes an economic decision.

Distractors:
\begin{enumerate}
\item when someone makes an economic request.
\item when someone makes an economic proposition.
\item when someone makes a political decision.
\end{enumerate}

\subsubsection*{Example 3}

Question: 
What did Chie hear? (SAT practice test 1 article 1)

Correct answer: 
her soft scuttling footsteps, the creak of the door.

Distractors:
\begin{enumerate}
\item her soft scuttling footsteps, the creak of the driveway.
\item her soft scuttling footsteps, the creak of the stairwell.
\item her soft scuttling footsteps, the knock of the door.
\end{enumerate}

\subsubsection*{Example 4}
Question: 
Who might duplicate itself? (SAT practice test 1 article 3)

Correct answer: 
the deoxyribonucleic acid molecule.

Distractors:
\begin{enumerate}
\item the deoxyribonucleic acid coenzyme.
\item the deoxyribonucleic acid polymer.
\item the deoxyribonucleic acid trimer
\end{enumerate}

\subsubsection*{Example 5}

Question: 
When does Deep Space Industries of Virginia hope to be harvesting metals from asteroids?
(SAT practice test 1 article 5)

Correct answer: 
by 2020.

Distractors:
\begin{enumerate}
\item by 2021.
\item by 2030.
\item by 2019.
\end{enumerate}


\subsubsection*{Example 6}


Question: 
What did a British study of the way women search for medical information online indicate?
(SAT practice test 2 article 3)

Correct answer: 
An experienced Internet user can, at least in some cases, assess the trustworthiness and probable value of a Web page in a matter of seconds.

Distractors:
\begin{enumerate}
\item An experienced Supernet user can, at least in some cases, assess the trustworthiness and probable value of a Web page in a matter of seconds.
\item An experienced CogNet user can, at least in some cases, assess the trustworthiness and probable value of a Web page in a matter of seconds.
\item An inexperienced Internet user can, at least in some cases, assess the trustworthiness and probable value of a Web page in a matter of seconds.
\end{enumerate}

\subsubsection*{Example 7}

What does a woman know better than a man? (SAT test 2 article 4)

Correct answer: 
the cost of life.

Distractors:
\begin{enumerate}
\item the cost of happiness.
\item the cost of experience.
\item the risk of life.
\end{enumerate}

\subsubsection*{Example 8}

This example presents a distractor without sufficient distraction.

Question: 
What are subject to egocentrism, social projection, and multiple attribution errors?

Correct answer: 
their insights.

Distractors:
\begin{enumerate}
\item their perspectives.
\item their findings.
\item their valuables. 
\end{enumerate}
The last distractor can be spotted wrong by just looking at the question: It is easy to tell
that it is out of place without the need to read the article.

\section{\uppercase{Conclusions and Final Remarks}}
\label{sec:conclusion}
\noindent
We presented a novel method using various NLP tools for generating adequate distractors for a QAP to form
an adequate MCQ on a given article. This is an interesting area with important applications. Experiments and evaluations on MCQs generated from the SAT 
practice reading tests indicate that our approach is promising. 

A number of improvements 
can be explored. For example, we may improve the ranking measure to help select a better
distractor for a target word from a list of candidates.
%
%
Another direction is 
explore how to produce generative distractors using neural networks, instead of just
replacing a few target words in a given answer. 

\section*{\uppercase{Acknowledgment}}
\noindent
This work was supported in part by funding from Eola Solutions, Inc. We thank
Hao Zhang and Changfeng Yu for discussions.

\bibliographystyle{apalike}
{\small
\bibliography{DG}}



\end{document}